\documentclass{article}
\PassOptionsToPackage{numbers}{natbib}
\usepackage[final]{nips_2017}

\usepackage[utf8]{inputenc} % allow utf-8 input
\usepackage[T1]{fontenc}    % use 8-bit T1 fonts
\usepackage{hyperref}       % hyperlinks
\usepackage{graphicx} % more modern
\usepackage{subfigure}

\usepackage{algorithm}
\usepackage{algorithmic}
\usepackage{amsfonts}
\usepackage{amsmath}
\usepackage{amssymb}
\usepackage[nohyperlinks,nolist]{acronym}
\usepackage{xparse}
\usepackage{wrapfig}

\usepackage{tikz}
\usepackage{array,colortbl,multirow,multicol,booktabs,ctable}
\usetikzlibrary{arrows}
\usetikzlibrary{positioning}
\usepackage{hyperref}
\usepackage[colorinlistoftodos,prependcaption,textsize=tiny]{todonotes}
\usepackage{wrapfig}
\definecolor{snsblue}{RGB}{76,114,176}
\definecolor{snsgreen}{RGB}{85,168,104}
\definecolor{snsred}{RGB}{197,78,82}
\definecolor{snspurple}{RGB}{130,114,179}
\definecolor{snsyellow}{RGB}{205,186,116}
\definecolor{snscyan}{RGB}{100,182,206}

\graphicspath{{plots/}}

\begin{acronym}
    \acro{RNN}[RNN]{recurrent neural network}
\end{acronym}

\newcommand{\model}[1]{\texttt{#1}}
\newcommand{\ets}{\model{ETS}}
\newcommand{\snyder}{\model{Snyder}}
\newcommand{\matfact}{\model{MatFact}}
\newcommand{\croston}{\model{Croston}}
\newcommand{\issm}{\model{ISSM}}
\newcommand{\issmfeat}{\model{ISSM}}
\newcommand{\gaussian}{\model{rnn-gaussian}} % gaussian, eg dividing all time-series by training target mean and multiply them for prediction, uniform sampling
\newcommand{\negbin}{\model{rnn-negbin}} % uniform sampling and no scaling
\newcommand{\deepar}{\model{DeepAR}}

\newcommand{\dataset}[1]{\texttt{#1}}
\newcommand{\parts}{\dataset{parts}}
\newcommand{\ec}{\dataset{ec}}
\newcommand{\ecsub}{\dataset{ec-sub}}
\newcommand{\electricity}{\dataset{electricity}}
\newcommand{\traffic}{\dataset{traffic}}

\newcommand{\z}[2]{z_{#1, #2}}
\newcommand{\xbf}{\mathbf{x}}
\newcommand{\zVec}[3]{\mathbf{z}_{#1, #2:#3}}
\newcommand{\xVec}[3]{\mathbf{x}_{#1, #2:#3}}

\newcommand{\powerlawFigure}{
\begin{wrapfigure}{R}{0.4\textwidth}
  \begin{center}
    \includegraphics[width=0.4\columnwidth]{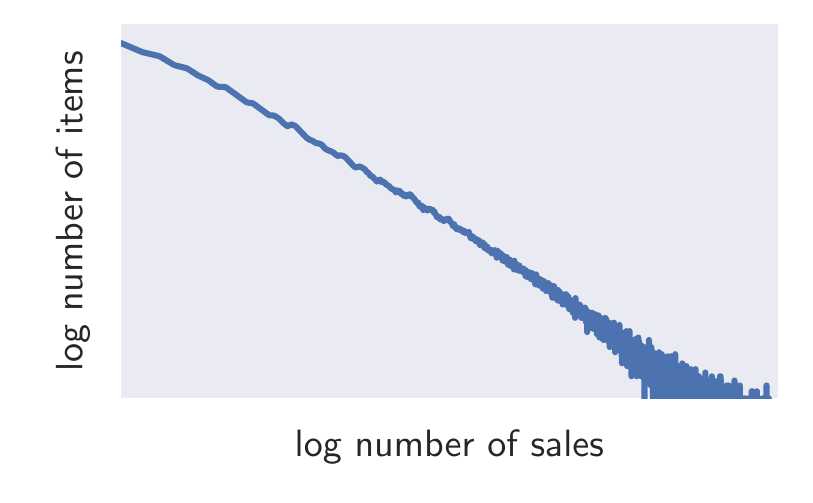}
  \end{center}
  \caption{
    Log-log histogram of the number of items versus number of sales for
    the 500K time series of \ec{}, showing the scale-free nature (approximately straight line) present in
    the \ec{} dataset (axis labels omitted due to the non-public nature of the data). 
    \label{fig:powerlaw}
}
\end{wrapfigure}
}

\NewDocumentCommand{\networkPicture}{s}{%
\begin{tikzpicture}[shorten >=1pt,->,draw=black!50, font=\small, scale=0.48]%
    \tikzstyle{every pin edge}=[<-,shorten <=1pt]
    \tikzstyle{node}=[circle,fill=gray!20,minimum size=25pt,inner sep=0pt]
    \tikzstyle{box}=[rectangle,draw=black!50,minimum height=12pt, inner sep=0pt]
    \tikzstyle{input node}=[box, minimum width=51pt,draw=black!30];
    \tikzstyle{network node}=[box, draw, fill=gray!20, minimum width=32pt];
    \tikzstyle{output node}=[box, minimum width=42pt,inner sep=2pt];
    \tikzstyle{sample node}=[node];
    \tikzstyle{node dots}=[node, fill=black, scale=0.2];
    \tikzstyle{annot} = [text width=4em, text centered]

     \foreach \i/\t/\tm/\tp in {0/{t-1}/{t-2}/{t},1/{t}/{t-1}/{t+1},2/{t+1}/{t}/{t+2}} {%
       \IfBooleanTF{#1}{
         \node[input node] (x\i) at (4*\i,0 ) {$\tilde{z}_{i,{\tm}}, x_{i, {\t}}$};%
       }{
         \node[input node] (x\i) at (4*\i,0 ) {$z_{i,{\tm}}, x_{i, {\t}}$};%
       }
       \node[network node] (f\i) at (4*\i,1.8) {$\hVec_{i, {\t}}$};%
       \node[output node] (y\i) at (4*\i,3.6) {$\ell(z_{i, \t}|\theta_{i, \t})$};%
       \IfBooleanTF{#1}{
         \node[sample node] (z\i) at (4*\i,5.8) {$\tilde{z}_{i, \t}$};
       }{
         \node[sample node] (z\i) at (4*\i,5.8) {$z_{i, \t}$};
       }
       \path (x\i) edge (f\i);%
       \path (f\i) edge (y\i);%
       \path (y\i) edge (z\i);%
     }%
     \node [left of=x0, node distance=0.62in, align=flush right] (xtlabel) {inputs};%
     \node [left of=f0, node distance=0.57in, align=flush right] (ftlabel) {network};%
     \IfBooleanT{#1}{
       \node [left of=z0, node distance=0.59in, align=flush right] (ytlabel) {samples};%
       \node [left of=z0, node distance=0.54in, align=flush right, shift={(0,-0.5)}] (ytlabel) {$\tilde{z} \sim \ell(\cdot|\theta)$};%
     }
     \draw (f0) edge (f1);%
     \draw (f1) edge (f2);%
       \IfBooleanT{#1}{
         \draw[dashed] (z0) edge (x1);%
         \draw[dashed] (z1) edge (x2);%
         }
\end{tikzpicture}%
}

\title{DeepAR: Probabilistic Forecasting with Autoregressive Recurrent Networks}

\author{
  David Salinas,  ~~Valentin Flunkert, ~~Jan Gasthaus \\
  Amazon Research\\
  Germany\\
  \texttt{<{dsalina,flunkert,gasthaus}@amazon.com>} 
}

\begin{document}

\maketitle

\begin{abstract}
Probabilistic forecasting, i.e.\ estimating the probability distribution
of a time series' future given its past, is a key enabler for optimizing business processes.
In retail businesses, for example, forecasting demand is crucial for having the right
inventory available at the right time at the right place.
In this paper we propose DeepAR, a methodology for producing accurate probabilistic forecasts, based on
training an auto-regressive recurrent network model on a large number of
related time series. We demonstrate how by applying deep learning techniques to forecasting,
one can overcome many of the challenges faced by widely-used classical approaches to the problem.
We show through extensive empirical evaluation on several real-world forecasting data sets accuracy improvements of around 15\% compared to state-of-the-art methods.
\end{abstract}

\section{Introduction}
\label{introduction}

Forecasting plays a key role in automating and optimizing
operational processes in most businesses and enables data driven decision making.
In retail for example, probabilistic forecasts of product supply and demand can be used for 
optimal inventory management, staff scheduling and topology planning~\cite{simchi2001},
and are more generally a crucial technology for most aspects of
supply chain optimization.

The prevalent forecasting methods in use today have been developed in the setting
of forecasting individual or small groups of time series. 
In this approach, 
model parameters for each given time series are independently estimated from past observations.
The model is typically manually selected to account for different
factors, such as autocorrelation structure, trend, seasonality, and other explanatory variables. 
The fitted model is then used to forecast the time series into the future
according to the model dynamics, possibly admitting probabilistic forecasts
through simulation or closed-form expressions for the predictive distributions.
Many methods in this class are based on the classical Box-Jenkins methodology \citep{box1968some},
exponential smoothing techniques, or state space models \citep{hyndman2008, seeger2016}.

In recent years, a new type of forecasting problem has become increasingly important in
many applications. Instead of needing to predict individual or a small number of time series,
one is faced with forecasting thousands or millions of related time series.
Examples include forecasting the energy consumption of
individual households, forecasting the load for servers in a data center, or
forecasting the demand for all products that a large retailer offers. In all these
scenarios, a substantial amount of data on past behavior of similar, related time series
can be leveraged for making a forecast for an individual time series. Using data from
related time series not only allows fitting more complex (and hence potentially more accurate)
models without overfitting, it can also alleviate the time and labor intensive manual 
feature engineering and model selection steps required by classical techniques.

In this work we present DeepAR, a forecasting method based on autoregressive
recurrent networks, which learns such a \emph{global} model from historical data
of \emph{all time series} in the data set. Our method builds upon previous work
on deep learning for time series data \cite{graves2013,
sutskever2014, wavenet}, and tailors a similar LSTM-based recurrent neural network
architecture to the probabilistic forecasting problem.

One challenge often encountered when attempting to jointly learn from multiple time
series in real-world forecasting problems is that the magnitudes of the time series differ widely, 
and the distribution of the magnitudes is strongly skewed.
This issue is illustrated in
Fig.~\ref{fig:powerlaw},
\powerlawFigure
which shows the distribution of sales
velocity (i.e.\ average weekly sales of an item) across millions of items sold by Amazon.
The distribution is over a few orders of magnitude an approximate
power-law.  This observation is to the best of our knowledge new (although
maybe not surprising) and has fundamental implications for forecasting methods
that attempt to learn global models from such datasets.  The scale-free nature of the distribution makes it difficult
to divide the data set into sub-groups of time series with a certain velocity band and learn separate models for them,
as each such velocity sub-group would have a similar skew.
Further, group-based regularization schemes, such as the one proposed by \citet{chapados2014}, may fail,
as the velocities will be vastly different within each group.
Finally, such skewed distributions make the use of certain commonly employed normalization techniques,
such input standardization or batch normalization \cite{BatchNorm}, less effective.
%The heavy tail implies that the network has to learn from examples in the tail and generalize to more extreme
%cases.  For instance, the model should be able to predict
%a time series with values that are larger than any example that it saw during training.

The main contributions of the paper are twofold: (1) we propose an RNN architecture for probabilistic forecasting,
incorporating a negative Binomial likelihood for count data as well as special treatment for the case when the magnitudes of the time series vary widely;
(2) we demonstrate empirically on several real-world data sets that this model produces accurate probabilistic forecasts across a range of input characteristics,
thus showing that modern deep learning-based approaches can effective address the probabilistic forecasting problem, 
which is in contrast to common belief in the field and the mixed results reported in \citep{ForecastingNNSOA, Kourentzes2013}.

In addition to providing better forecast accuracy than previous methods,
our approach has a number key advantages compared to classical approaches and other global methods: 
(i) As the model learns seasonal behavior and dependencies on given covariates across time series,
 minimal manual feature engineering is needed to capture complex, group-dependent behavior;
(ii) DeepAR
makes probabilistic forecasts in the form of Monte Carlo samples that can be used to compute consistent quantile estimates for all sub-ranges in the prediction horizon;
(iii) By learning from similar items, our method is able to provide forecasts for items with little or no history
at all, a case where traditional single-item forecasting methods fail;
(vi) Our approach does not assume Gaussian noise, but can incorporate a wide range of likelihood functions, allowing the user to choose one that is appropriate for the statistical properties of the data.

Points (i) and (iii) are what set DeepAR apart from classical forecasting approaches, 
while (ii) and (iv) pertain to producing
accurate, calibrated forecast distributions learned from the historical
behavior of all of the time series jointly, which is not addressed by other global methods (see Sec. \ref{sec:relatedWork}).
Such probabilistic forecasts are of crucial importance in many applications, as they---in contrast to point forecasts---enable optimal decision making
under uncertainty by minimizing risk functions, i.e.\ expectations of some loss function under the forecast distribution.

\section{Related Work}
\label{sec:relatedWork}

Due to the immense practical importance of forecasting, a vast variety of
different forecasting methods have been developed. Prominent examples of methods
for forecasting individual time series include
ARIMA models \cite{box1968some} and exponential smoothing methods;
\citet{hyndman2008} provide a unifying review of these and related techniques.
 
Especially in the demand forecasting domain, one is often faced with highly erratic, intermittent
or bursty data which violate core assumptions of many classical techniques, such as
Gaussian errors, stationarity, or homoscedasticity of the time series. Since data preprocessing methods~(e.g.\ \cite{boxcox1964})
often do not alleviate these conditions, forecasting methods have also incorporated more suitable likelihood functions, such as
the zero-inflated Poisson distribution, the negative binomial distribution \cite{snyder2012},
a combination of both \cite{chapados2014}, or a tailored multi-stage likelihood
\cite{seeger2016}.
 
Sharing information across time series can improve the forecast accuracy, 
but is difficult to accomplish in practice, because of the often heterogeneous nature of the
data.  Matrix factorization methods (e.g.\ the recent work of
\citet{NIPS2016_6160}), as well as Bayesian methods that share information via
hierarchical priors \cite{chapados2014} have been proposed as 
mechanisms for
learning across multiple related time series and leveraging hierarchical structure~\cite{hyndman2011}.

%In contrast to existing work, our approach allows  full flexibility in the use of likelihoods. Any of the afore-mentioned 
%likelihoods can for instance be used. Information sharing is accomplished by a neural network that is trained 
%on all time-series.

Neural networks have been investigated in the context of forecasting for a long
time (see e.g.\ the numerous references in the survey \cite{ForecastingNNSOA},
or \cite{Gers2001} for more recent work considering LSTM cells).  More
recently, \citet{Kourentzes2013} applied neural networks specifically to
intermittent data but obtained mixed results.  Neural networks in forecasting
have been typically applied to individual time series, i.e.\ a different model
is fitted to each time series independently
\cite{kaastra1996designing,ghiassi2005dynamic,diaz2008hybrid}. On the other
hand, outside of the forecasting community, time series models based on
recurrent neural networks have been very successfully applied to other
applications, such as natural language processing \cite{graves2013,
sutskever2014}, audio modeling \cite{wavenet} or image generation
\cite{gregor2015draw}.  Two main characteristics make the forecasting
setting that we consider here different: First, in probabilistic forecasting one is
interested in the full predictive distribution, not just a single best
realization, to be used in downstream decision making systems. Second, to obtain
accurate distributions for (unbounded) count data, we use a negative Binomial likelihood,
which improves accuracy but precludes us from directly applying standard data normalization techniques.

\newcommand{\hVec}{\mathbf{h}}

\section{Model}
\newcommand{\modelFigure}{
\begin{figure}[t]
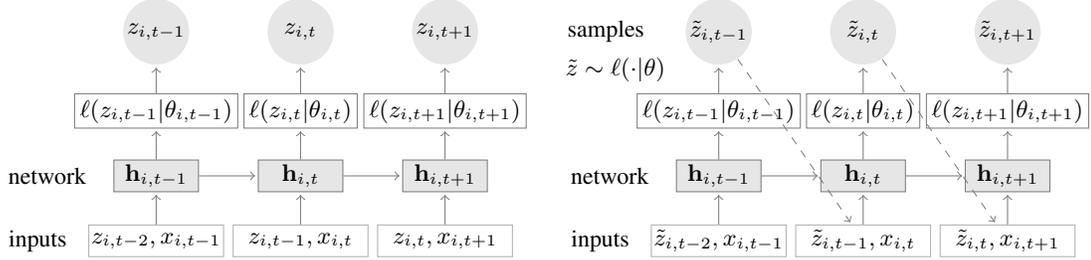

\centering
\begin{center}
    \networkPicture~~~~~~\networkPicture*
\end{center}
\caption{
    Summary of the model. Training (left): At each time step $t$, the inputs to the network are the covariates $x_{i,t}$,
    the target value at the previous time step $z_{i,t-1}$, as well as the previous network output $\hVec_{i, t-1}$.
    The network output $\hVec_{i,t} = h(\hVec_{i,t-1}, z_{i,t-1}, \mathbf{x}_{i,t}, \Theta)$ is then used to compute the parameters $\theta_{i,t} = \theta(\hVec_{i,t}, \Theta)$ of the likelihood $\ell(z|\theta)$,
    which is used for training the model parameters. For prediction,
    the history of the time series $z_{i,t}$ is fed in for $t<t_0$, then in the prediction range (right)
    for $t\ge t_0$ a sample
    $\hat{z}_{i,t} \sim \ell(\cdot|\theta_{i,t})$ is drawn and fed back for the next point 
    until the end of the prediction range $t=t_0 + T$ generating one sample trace.
    Repeating this prediction process yields many traces representing the joint predicted distribution.
   \label{fig:encoderdecoder}}
\end{figure}
}

\modelFigure

Denoting the value of time series $i$ at time $t$ by $\z{i}{t}$,
our goal is to model the conditional distribution
\begin{equation*}
    P(\zVec{i}{t_0}{T} | \zVec{i}{1}{t_0-1}, \xVec{i}{1}{T})
    \label{eq:condDist}
\end{equation*}
of the future of each
time series $[\z{i}{t_0}, \z{i}{t_0 + 1}, \ldots, \z{i}{T}] := \zVec{i}{t_0}{T}$ given its 
\hbox{past $[\z{i}{1}, \ldots, \z{i}{t_0-2}, \z{i}{t_0-1}] := \zVec{i}{1}{t_0-1}$},
where $t_0$ denotes the time point from which we assume $\z{i}{t}$ to be unknown at prediction time,
and $\xVec{i}{1}{T}$ are covariates that are assumed to be known for all time points. To prevent
confusion we avoid the ambiguous terms ``past'' and ``future'' and will refer to time ranges $[1, t_0-1]$ and $[t_0, T]$ as the conditioning range and 
prediction range, respectively. During training, both ranges have to lie in the past so that the $\z{i}{t}$ are observed, but during prediction $\z{i}{t}$
is only available in the conditioning range. Note that the time index $t$ is relative, i.e.\ $t=1$ can correspond to a different actual
time period for each $i$. 

\newcommand{\modelDist}{Q_\Theta(\zVec{i}{t_0}{T} | \zVec{i}{1}{t_0-1}, \xVec{i}{1}{T})}

Our model, summarized in Fig.~\ref{fig:encoderdecoder}, is based on an autoregressive recurrent network
architecture \cite{graves2013,sutskever2014}.
We assume that our model distribution $\modelDist$
consists of a product of likelihood factors
\begin{align*}
    \modelDist &= \prod\nolimits_{t=t_0}^T Q_\Theta(z_{i,t}|\mathbf{z}_{i,1:t-1}, \xVec{i}{1}{T}) = \prod\nolimits_{t=t_0}^T \ell(\z{i}{t} | \theta(\hVec_{i, t}, \Theta))
\end{align*}
 parametrized by the output $\hVec_{i, t}$ of an autoregressive recurrent network
\begin{equation}
    \hVec_{i, t} = h\left(\hVec_{i, t-1}, \z{i}{t-1}, \xbf_{i, t}, \Theta\right) \,,
    \label{eq:recurrence}
\end{equation}
where $h$ is a function implemented by a multi-layer recurrent neural network with LSTM cells.% 
\footnote{Details of the architecture and hyper-parameters are given in the supplementary material.}
The model is autoregressive, in the sense that it consumes the observation at the last time step $\z{i}{t-1}$ as an input,
as well as recurrent, i.e.\ the previous output of the network $\hVec_{i,t-1}$ is fed back as an input at the next time step.
The likelihood $\ell(\z{i}{t}|\theta(\hVec_{i,t}))$ is a fixed distribution
whose parameters are given by a function $\theta(\hVec_{i,t}, \Theta)$ of the network output $\hVec_{i, t}$ (see below).

Information about the observations in the conditioning range $\zVec{i}{1}{t_0 -1}$ is transferred to the
prediction range through the initial state $\hVec_{i, t_0-1}$. In the sequence-to-sequence setup, this initial state is
 the output of an \emph{encoder network}. While in general this encoder network can have a different architecture, in our 
 experiments we opt for using the
 same architecture for the model in the conditioning range and the prediction range (corresponding to the \emph{encoder} and \emph{decoder} in
 a sequence-to-sequence model). Further, we share weights between them, so that the initial state
 for the decoder $\hVec_{i, t_0 - 1}$ is
obtained by computing \eqref{eq:recurrence} for $t = 1, \ldots, t_0 - 1$, where all required quantities are observed.
The initial state of the encoder $\hVec_{i, 0}$ as well as $\z{i}{0}$ are initialized to zero.

Given the model parameters $\Theta$, we can directly obtain joint samples
$\tilde{\mathbf{z}}_{i, t_0:T} \sim \modelDist$ through ancestral sampling:
First, we obtain $\hVec_{i, t_0-1}$ by computing \eqref{eq:recurrence} for $t=1,\ldots, t_0$.
For $t=t_0, t_0+1, \ldots, T$ we sample $\tilde{z}_{i, t} \sim \ell(\cdot | \theta(\tilde{\mathbf{h}}_{i,t}, \Theta))$
where $\tilde{\mathbf{h}}_{i, t} = h\left(\hVec_{i, t-1}, \tilde{z}_{i, t-1}, \xbf_{i, t}, \Theta\right)$
initialized with $\tilde{\mathbf{h}}_{i, t_0-1} = \hVec_{i, t_0-1}$ and $\tilde{z}_{i, t_0 -1} = \z{i}{t_0 - 1}$.
Samples from the model obtained in this way can then be used to compute quantities
of interest, e.g.\ quantiles of the distribution of the sum of values for some
time range in the future.

\subsection{Likelihood model}
\label{sec:likelihood}

The likelihood $\ell(z|\theta)$ determines the ``noise model'', and should be
chosen to match the statistical properties of the data. In our approach, the
network directly predicts \emph{all} parameters $\theta$ (e.g.\ mean \emph{and} variance) of the
probability distribution for the next time point.

For the experiments in this paper, we consider two choices, Gaussian likelihood for real-valued data, and negative-binomial likelihood for positive count data.
Other likelihood models can also readily be used, e.g.\ beta likelihood for data in the unit interval, Bernoulli
likelihood for binary data, or mixtures in order to handle complex marginal distributions, as long as samples from the distribution can cheaply be obtained, and the log-likelihood and its gradients wrt.\ the parameters can be evaluated.
We parametrize the Gaussian likelihood using its mean and standard deviation, $\theta = (\mu, \sigma)$, where the mean is given by an affine function of the network output, and the standard deviation
is obtained by applying an affine transformation followed by a softplus activation in order to ensure $\sigma > 0$:
\begin{align*}
    \ell_\text{G}(z|\mu, \sigma) &= (2\pi\sigma^2)^{-\frac{1}{2}}\exp(-(z - \mu)^2/(2\sigma^2))\\
    \mu(\hVec_{i, t}) &= \mathbf{w}_\mu^T \hVec_{i, t} + b_\mu \quad\text{and}\quad \sigma(\hVec_{i, t}) = \log(1 + \exp(\mathbf{w}_\sigma^T \hVec_{i, t} + b_\sigma)) \;.
\end{align*}
For modeling time series of positive count data, the negative binomial distribution is a commonly
used choice \cite{snyder2012, chapados2014}.
We parameterize the negative binomial distribution
by its mean $\mu \in \mathbb{R}^+$ and a shape parameter $\alpha \in \mathbb{R}^+$,
\begin{align*}
    \ell_{\text{NB}}(z|\mu, \alpha) &= \frac{\Gamma(z + \frac{1}{\alpha})}{\Gamma(z + 1)\Gamma(\frac{1}{\alpha})} \left(\frac{1}{1 + \alpha \mu}\right)^\frac{1}{\alpha} \left(\frac{\alpha \mu}{1 + \alpha \mu}\right)^z \\
    \mu(\hVec_{i, t}) &= \log(1 + \exp(\mathbf{w}_\mu^T \hVec_{i, t} + b_\mu)) \quad\text{and}\quad \alpha(\hVec_{i, t}) = \log(1 + \exp(\mathbf{w}_\alpha^T \hVec_{i, t} + b_\alpha)) \;,
\end{align*}
where both parameters are obtained from the network output by a fully-connected layer with softplus activation to ensure positivity.
In this parameterization of the negative binomial distribution the shape parameter $\alpha$ scales the variance relative to the mean, i.e.\ $\text{Var}[z] = \mu + \mu^2 \alpha$.
While other parameterizations are possible, we found this particular one to be especially conducive to fast convergence in preliminary experiments.

\subsection{Training}
Given a data set of time series $\{\zVec{i}{1}{T}\}_{i = 1, \ldots, N}$ and associated covariates $\xVec{i}{1}{T}$,
obtained by choosing a time range such that $\z{i}{t}$ in the prediction range is known, the parameters $\Theta$ of the model,
consisting of the parameters of the RNN $h(\cdot)$ as well as the parameters of $\theta(\cdot)$,
can be learned by maximizing the \hbox{log-likelihood}
\begin{equation}
    \mathcal{L} = \sum_{i=1}^{N} \sum_{t=t_0}^{T} \log \ell(\z{i}{t}|\theta(\hVec_{i, t}))\,.
    \label{eq:KL}
\end{equation}
As $\hVec_{i,t}$ is a deterministic function of the input,
all quantities required to compute \eqref{eq:KL} are observed, so that---in
contrast to state space models with latent variables---no inference is
required, and \eqref{eq:KL} can be optimized directly via stochastic gradient descent
by computing gradients with respect to $\Theta$.
In our experiments, where the encoder model is the same as the decoder, the
distinction between encoder and decoder is somewhat artificial during
training, so that we also include the likelihood terms
for $t=0, \ldots, t_0 - 1$ in \eqref{eq:KL} (or, equivalently, set $t_0 =
0$).

%A forecast that only predicts independent marginal
%distributions for each time-point can already generate fairly accurate
%forecasts\footnote{Plots such as Fig.~\ref{fig:predictionsplots} show marginal distributions i.e.\ correlated/uncorrelated samples are indistinguishable.}
%and while there is usually a small positive correlation that needs to be
%captured to achieve high accuracy (see Fig.~\ref{fig:calibration} and
%discussion), the correlation is typically much less important than in other
%problem domains. In a machine translation task, for instance, predicting
%independent distributions for each word in a sentence would result in a very
%bad model, because the next word highly depends on the previous word. 

For each time series in the dataset, we generate multiple training instances
by selecting windows with different starting points from the original
time series. In practice, we keep the total length $T$ as well as the relative
length of the conditioning and prediction ranges fixed for all training examples.
For example, if the total available range for a given time series ranges
from 2013-01-01 to 2017-01-01, we can create training examples
with $t=1$ corresponding to 2013-01-01, 2013-01-02, 2013-01-03, and so on.
When choosing these windows we ensure that entire prediction range is always covered
by the available ground truth data, but we may chose $t=1$ to lie \emph{before}
the start of the time series, e.g.\ 2012-12-01 in the example above, padding
the unobserved target with zeros. This allows the model to learn the behavior of
``new'' time series taking into account all other available features.
By augmenting the data using this windowing procedure, we ensure that information about
absolute time is only available to the model through covariates, but not through the
relative position of $z_{i,t}$ in the time series.

\citet{bengio2015} noted that, due to the autoregressive nature of such models,
optimizing \eqref{eq:KL} directly causes a discrepancy between how the model is
used during training and when obtaining predictions from the model: during training,
the values of $\z{i}{t}$ are known in the prediction range and can be used to
compute $\hVec_{i, t}$; during prediction however, $\z{i}{t}$ is unknown for $t
\geq t_0$, and a single sample $\tilde{z}_{i, t} \sim \ell( \cdot| \theta(\hVec_{i,
t}))$ from the model distribution is used in the computation of $\hVec_{i, t}$
according to \eqref{eq:recurrence} instead. While it has been shown that 
this disconnect poses a severe problem for e.g.\ NLP tasks, we have not observed 
adverse effects from this in the forecasting setting.
Preliminary experiments with variants of scheduled sampling
\citep{bengio2015} did not show any significant accuracy improvements (but slowed convergence).

\subsection{Scale handling} Applying the model to data that exhibits
a power-law of scales as depicted in Fig.~\ref{fig:powerlaw} presents two
challenges.  Firstly, due to the autoregressive nature of the model, both the
autoregressive input $z_{i,t-1}$ as well as the output of the network (e.g.\
$\mu$) directly scale with the observations $z_{i,t}$, but the non-linearities
of the network in between have a limited operating range.  Without further
modifications, the network thus has to learn to scale the input to an
appropriate range in the input layer, and then to invert this scaling at the
output. We address this issue by dividing the autoregressive inputs $z_{i,t}$
(or $\tilde{z}_{i,t}$) by an item-dependent scale factor $\nu_i$, and
conversely multiplying the scale-dependent likelihood parameters by the same
factor. For instance, for the negative binomial likelihood we use $\mu=\nu_i
\log(1 + \exp(o_\mu))$ and $\alpha = \log(1 + \exp(o_\alpha))/\sqrt{\nu_i}$ 
where $o_\mu$, $o_\alpha$ are the outputs of
the network for these parameters. Note that while for real-valued data one could alternatively
scale the input in a preprocessing step, this is not possible for count distributions.
Choosing an appropriate scale factor might in itself be challenging (especially
in the presence of missing data or large within-item variances).
However, scaling by the average value $\nu_i = 1 + \frac{1}{t_0} \sum_{t=1}^{t_0} z_{i,t}$,
as we do in our experiments, is a heuristic that works well in practice.

Secondly, due to the imbalance in the data, a stochastic optimization procedure
that picks training instances uniformly at random will visit the small number
time series with a large scale very infrequently, which result in underfitting
those time series.  This could be especially problematic in the demand
forecasting setting, where high-velocity items can exhibit qualitatively
different behavior than low-velocity items, and having an accurate forecast for
high-velocity items might be more important for meeting certain business
objectives.  To counteract this effect, we sample the examples non-uniformly
during training. In particular, in our weighted sampling scheme, the probability of selecting
a window from an example with scale $\nu_i$ is proportional to $\nu_i$. This
sampling scheme is simple, yet effectively compensates for the skew in Fig.~\ref{fig:powerlaw}.

\subsection{Features}
The covariates $\mathbf{x}_{i,t}$ can be item-dependent, time-dependent, or both.%
\footnote{Covariates $\mathbf{x}_{i,t}$ that do not depend on time are handled by repeating 
them along the time dimension.}
They can be used to provide additional information about the
item or the time point (e.g.\ week of year) to the model.
They can also be used to include covariates that one expects to influence the
outcome (e.g.\ price or promotion status in the demand forecasting setting),
as long as the features' values are available also in the prediction range.
\begin{figure*}
\center
% other plots are present in the folder in case you prefer to choose other ones
\includegraphics[width=.23\textwidth]{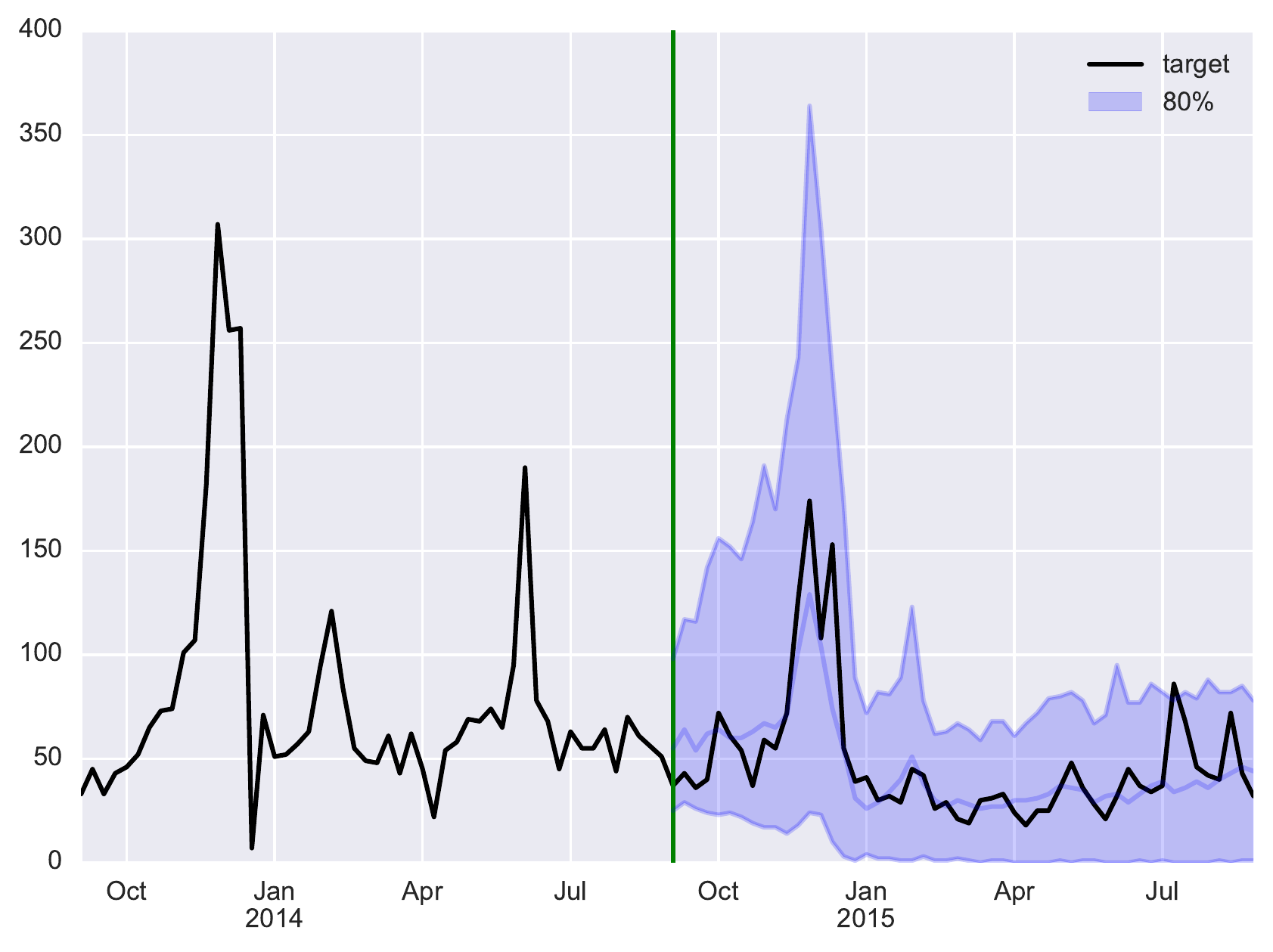} % fast & seasonal
\includegraphics[width=.23\textwidth]{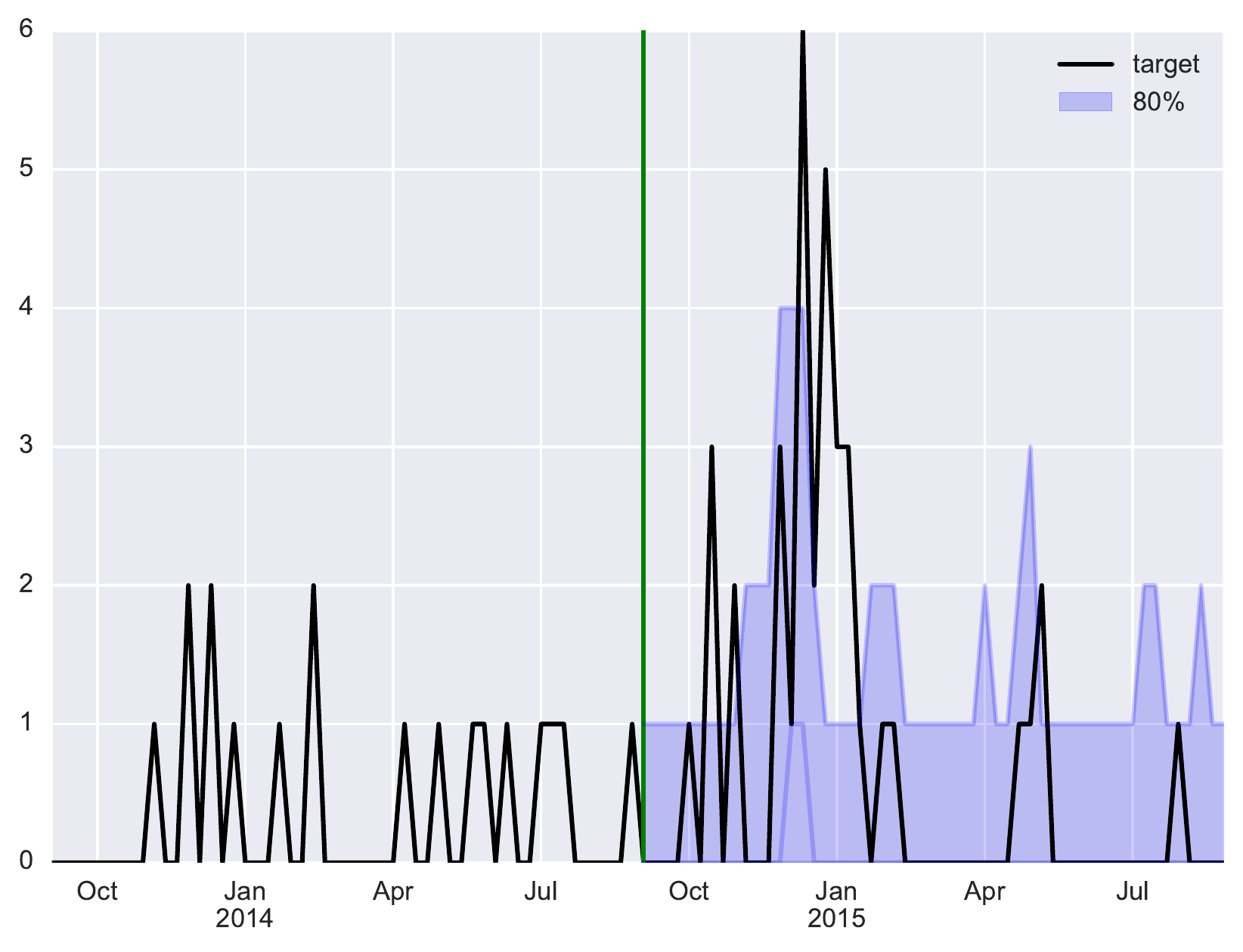} % slow & seasonal
\includegraphics[width=.23\textwidth]{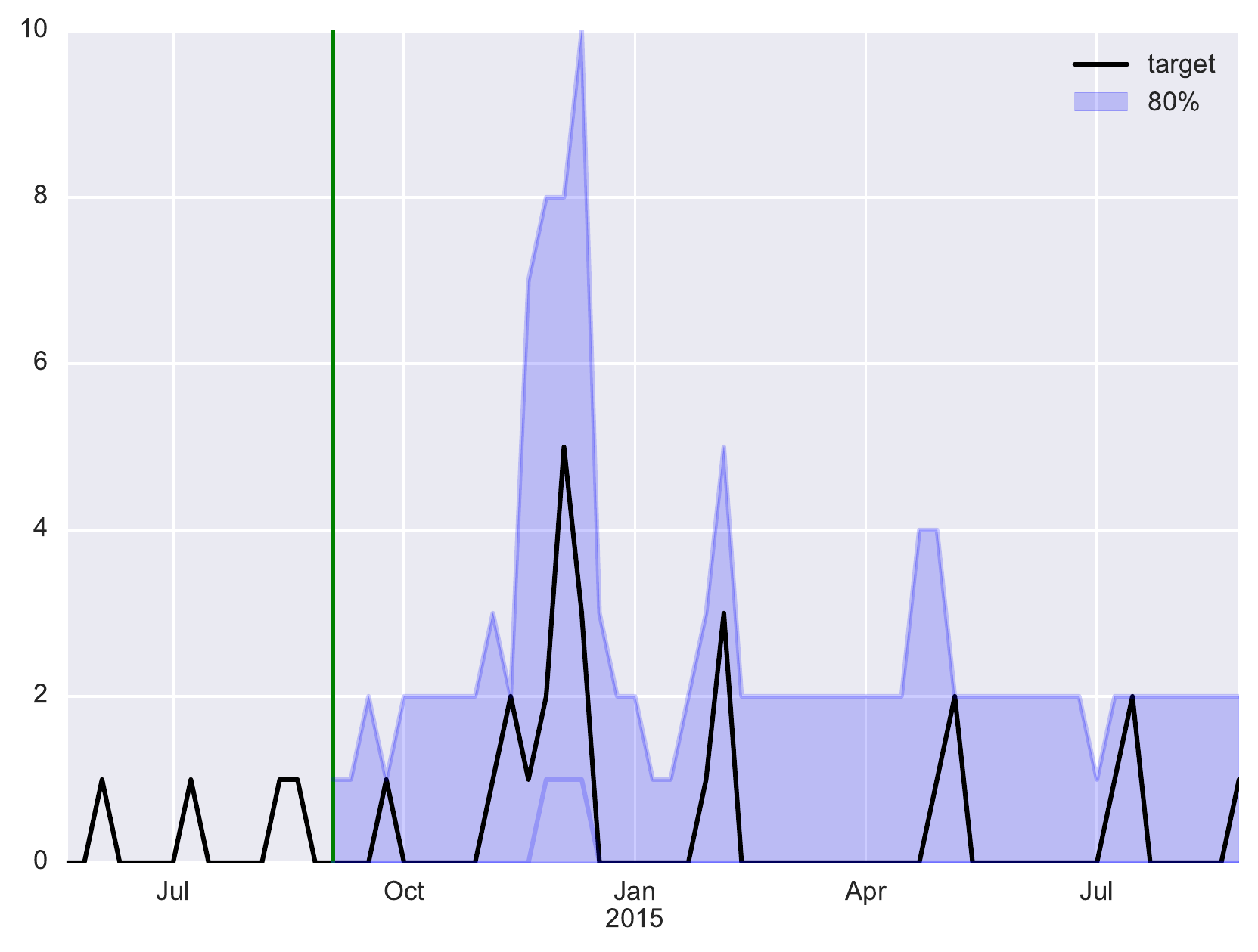} % new
\includegraphics[width=.23\textwidth]{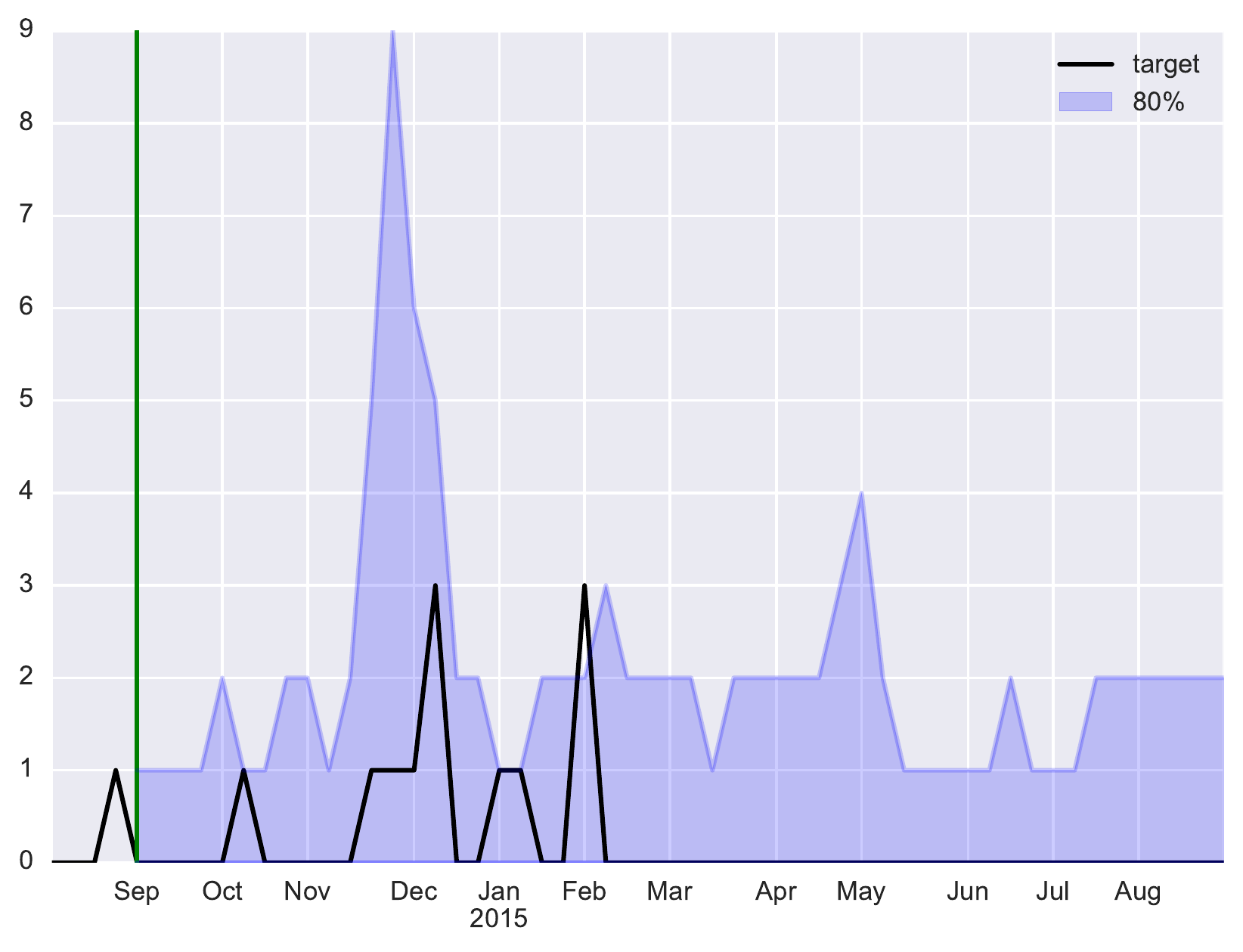} % new
\caption{Example time series of \ec{}. The vertical line
    separates the conditioning period from the prediction period.
    The black line shows the true target. In the prediction range
    we plot the p50 as a blue line (mostly zero for the three slow items)
    and the 80\% confidence interval (shaded).
    The model learns accurate seasonality patterns and uncertainty estimates
    for items of different velocity and age.
    \label{fig:predictionsplots}}
\end{figure*}
In all experiments we use an ``age'' feature, i.e., the distance to the first
observation in that time series.
We also add day-of-the-week and hour-of-the-day for hourly data,
week-of-year for weekly data and month-of-year for monthly data.%
\footnote{Instead of using dummy variables to encode these, we simply encode them
as increasing numeric values.}
Further, we include a single categorical item feature, for which an embedding
is learned by the model. In the retail demand forecasting data sets, the
item feature corresponds to a (coarse) product category (e.g.\ ``clothing''),
while in the smaller data sets it corresponds to the item's identity,
allowing the model to learn item-specific behavior.
We standardize all covariates to have zero mean and unit variance.

\section{Applications and Experiments}

We implement our model using MXNet, and use a single p2.xlarge AWS instance containing 4~CPUs
and 1~GPU to run all experiments. 
On this hardware, a full training \& prediction run on the large \ec{} dataset containing 500K
time series can be completed in less than 10 hours. While prediction is already fast, is can easily parallelized if necessary.
A description of the (simple) hyper-parameter tuning procedure, the obtained hyper-parameter values, as well as statistics of datasets and running time are given in
supplementary material.  

\textbf{Datasets} --
We use five datasets for our evaluations. The first three--\parts{}, \electricity{}, and \traffic{}--are public
datasets; \parts\ consists of 1046 aligned time series of 50 time steps each,
 representing monthly sales for different items of a US automobile company
 \cite{seeger2016}; 
 \electricity{} contains hourly time series of the electricity consumption of 370 customers \cite{NIPS2016_6160};
 \traffic{}, also used in \cite{NIPS2016_6160}, contains the hourly occupancy rate, between 0 and 1, of 963 car lanes of San Francisco bay area freeways.
For the \parts{} dataset, we use the 42 first months as training data and report error on the remaining 8. For \electricity{} we train with data between 2014-01-01 and 2014-09-01, for \traffic{} we train all the data available before 2008-06-15. The results for \electricity{} and \traffic{} are computed using rolling window predictions done after the last point seen in training as described in \cite{NIPS2016_6160}.
We do not retrain our model for each window, but use a single model trained on the data before the first prediction window.
The remaining two datasets \ec{} and \ecsub{} are weekly item sales from Amazon used in \cite{seeger2016}.
We predict 52 weeks and evaluation is done on the year following \texttt{2014-09-07}.
The time series in these two datasets are very diverse and erratic, ranging
from very fast to very slow moving items, and contains ``new'' products introduced
in the weeks before the forecast time \texttt{2014-09-07}, see
Fig.~\ref{fig:predictionsplots}. Further, item velocities in this data set have
a power-law distribution, as shown in Fig.~\ref{fig:powerlaw}.

\begin{figure}
\begin{minipage}[b]{0.5\linewidth}
    \centering
\includegraphics[width=\textwidth]{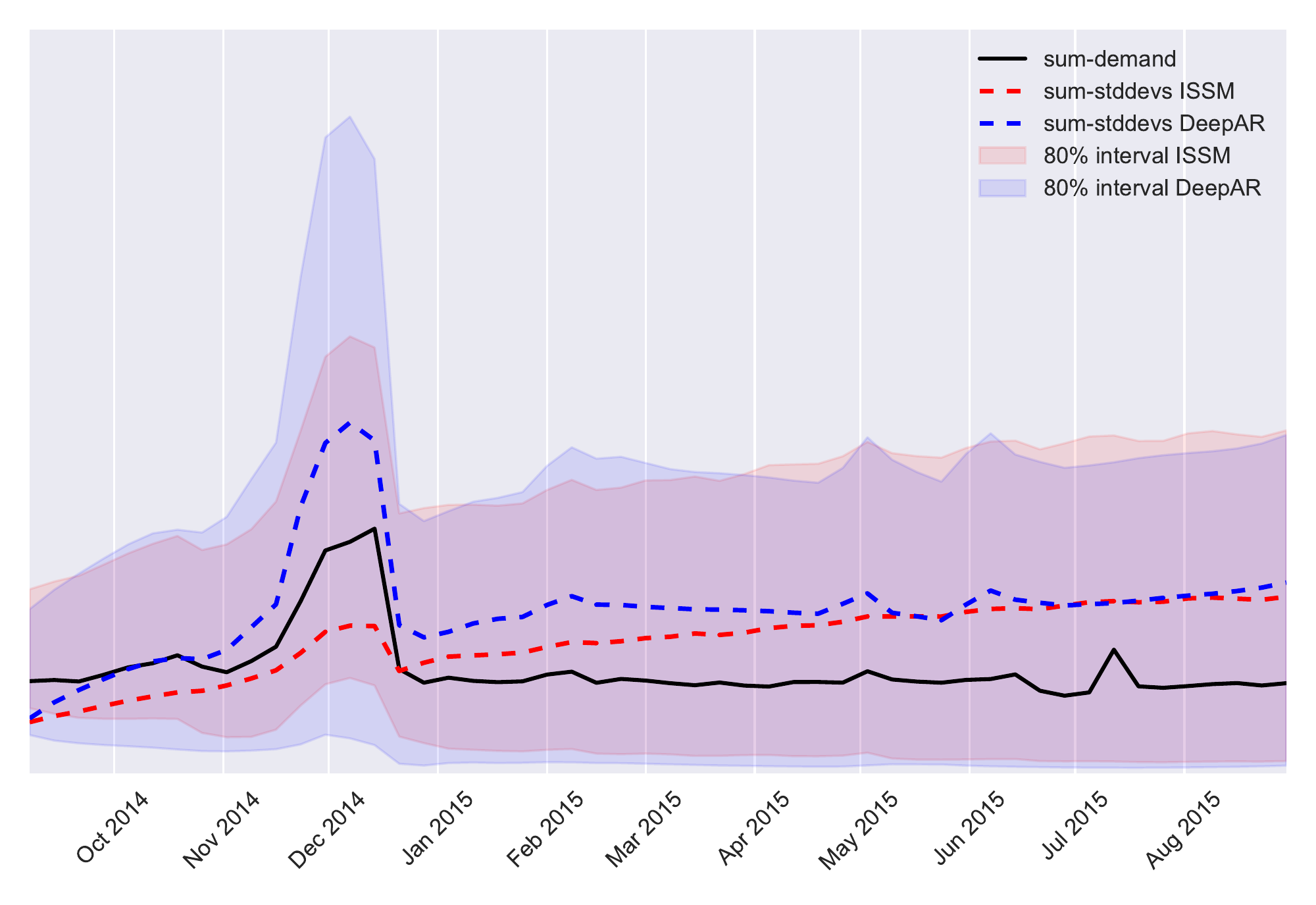}
\caption{Uncertainty growth over time for ISSM and DeepAR models.
Unlike the ISSM, which postulates a linear growth of uncertainty, the
behavior of uncertainty is learned from the data,
resulting in a non-linear growth with a (plausibly) higher uncertainty around Q4.
The aggregate is calculated over the entire \ec{} dataset. \label{fig:uncertainty}}
\end{minipage}
    \hspace{0.2cm}
\begin{minipage}[b]{0.5\linewidth}
\centering
\includegraphics[width=\textwidth]{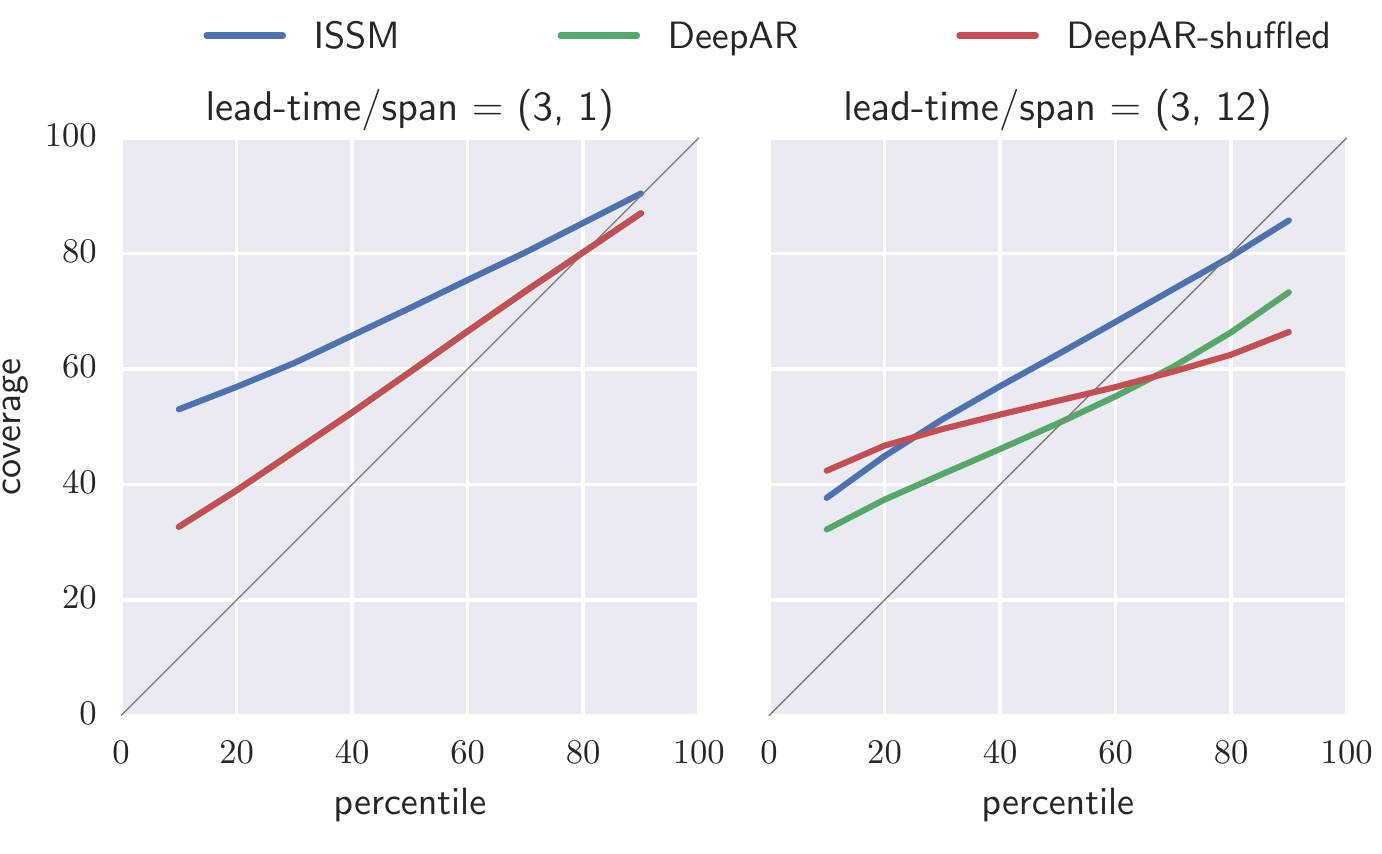}
\caption{Coverage for two spans on the \ecsub{} dataset. The left panel shows the coverage for a single time-step interval, while the right panel shows these metrics for a larger time interval with 9 time-steps. When correlation in the prediction sample paths is destroyed by shuffling the samples for each time step, correlation is destroyed and the forecast becomes less calibrated. This shuffled prediction also has a 10\% higher $0.9$-risk.
\label{fig:calibration}}
\end{minipage}
\end{figure}

\begin{table*}[t]
\scriptsize
\centering
\begin{tabular}{l|rrrrrrrrr}
\toprule
{} & \multicolumn{4}{c}{$0.5$-risk} & \multicolumn{4}{c}{$0.9$-risk} & average \\
\midrule
{} & \multicolumn{9}{c}{\parts} \\
$(L,S)$ & $(0,1)$ & $(2,1)$ & $(0,8)$ & all(8) & $(0,1)$ & $(2,1)$ & $(0,8)$ & all(8) & average\\
\midrule
\snyder{} (baseline) & 1.00 & 1.00 & 1.00 & \bf{1.00} & 1.00 & 1.00 & 1.00 & 1.00 & 1.00 \\
\croston{}  & 1.47 & 1.70 & 2.86 & 1.83 & - & - & - & - & 1.97 \\
\issmfeat{} & 1.04 & 1.07 & 1.24 & 1.06 & 1.01 & 1.03 & 1.14 & 1.06 & 1.08 \\
\ets{} & 1.28 & 1.33 & 1.42 & 1.38 & 1.01 & 1.03 & 1.35 & 1.04 & 1.23 \\
\gaussian{} & 1.17 & 1.49 & 1.15 & 1.56 & 1.02 & 0.98 & 1.12 & 1.04 & 1.19 \\
\negbin{} & \bf{0.95} & \bf{0.91} & 0.95 & \bf{1.00} & 1.10 & \bf{0.95} & 1.06 & 0.99 & 0.99 \\
% note results are a bit better than icml here because of the weighted sampling
\deepar{} & 0.98 & \bf{0.91} & \bf{0.91} & 1.01 & \bf{0.90} & \bf{0.95} & \bf{0.96} & \bf{0.94} & \bf{0.94} \\
\midrule
{} & \multicolumn{9}{c}{\ecsub} \\
    $(L,S)$ & $(0,2)$ & $(0,8)$ & (3, 12) & all(33) & $(0,2)$ & $(0,8)$ & (3, 12) & all(33) & average\\
\midrule
\snyder{}  & 1.04 & 1.18 & 1.18 & 1.07 & 1.0 & 1.25 & 1.37 & 1.17 & 1.16 \\
\croston{}  & 1.29 & 1.36 & 1.26 & 0.88 & - & - & - & - & 1.20 \\
\issmfeat{} (baseline) & 1.00 & 1.00 & 1.00 & 1.00 & 1.00 & 1.00 & \bf{1.00} & 1.00 & 1.00 \\
\ets{} & 0.83 & 1.06 & 1.15 & 0.84 & 1.09 & 1.38 & 1.45 & 0.74 & 1.07 \\
\gaussian{} & 1.03 & 1.19 & 1.24 & 0.85 & 0.91 & 1.74 & 2.09 & 0.67 & 1.21\\
\negbin{} & 0.90 & 0.98 & 1.11 & 0.85 & 1.23 & 1.67 & 1.83 & 0.78 & 1.17 \\
\deepar{} & \bf{0.64} & \bf{0.74} & \bf{0.93} & \bf{0.73} & \bf{0.71} & \bf{0.81} & 1.03 & \bf{0.57} & \bf{0.77} \\
\midrule
{} & \multicolumn{9}{c}{\ec} \\
$(L,S)$ & $(0,2)$ & $(0,8)$ & (3, 12) & all(33) & $(0,2)$ & $(0,8)$ & (3, 12) & all(33) & average\\
\midrule
\snyder{}  &  0.87 & 1.06 & 1.16 & 1.12 & 0.94 & 1.09 & 1.13 & 1.01 & 1.05 \\
\croston{}  & 1.30 & 1.38 & 1.28 & 1.39 & - & - & - & - & 1.34 \\
\issmfeat{} (baseline) & 1.00 & 1.00 & 1.00 & 1.00 & 1.00 & 1.00 & \bf{1.00} & 1.00 & 1.00 \\
\ets{} & 0.77 & 0.97 & 1.07 & 1.23 & 1.05 & 1.33 & 1.37 & 1.11 & 1.11 \\
\gaussian{} & 0.89 & 0.91 & 0.94 & 1.14 & 0.90 & 1.15 & 1.23 & \bf{0.90} & 1.01  \\
\negbin{} & 0.66 & 0.71 & \bf{0.86} & \bf{0.92} & 0.85 & 1.12 & 1.33 & 0.98 & 0.93 \\
\deepar{} &  \bf{0.59} & \bf{0.68} & 0.99 & 0.98 & \bf{0.76} & \bf{0.88} & \bf{1.00} & 0.91 & \bf{0.85} \\
\bottomrule
\end{tabular}
    \caption{Accuracy metrics relative to the strongest previously published method (baseline). Best results are marked in bold (lower is better). \label{tab:accuracy-distribution}}
\end{table*}

\subsection{Accuracy comparison}

For the \parts{} and \ec{}/\ecsub{} datasets we compare with the following baselines which represent the state-of-the-art on demand integer datasets to the best of our knowledge:

\begin{itemize}
\item \croston{}: the Croston method developed for intermittent demand forecasting from R package \citep{Rpackage}
\item \ets{}: the ETS model \citep{hyndman2008} from R package with automatic model selection. Only additive models are used as multiplicative models shows numerical issues on some time series.
\item \snyder{} the negative-binomial autoregressive method of \citep{snyder2012}
\item \issmfeat{} the method of \citep{seeger2016} using an innovative state space model with covariates features
\end{itemize}

In addition, we compare to two baseline RNN models to see the effect of our contributions: 

\begin{itemize}
\item \gaussian{} uses the same architecture
as \deepar{} with a Gaussian likelihood; however, it uses uniform sampling, and a simpler scaling mechanism, where time series $z_i$ are divided by $\nu_i$ and outputs are
multiplied by $\nu_i$
\item \negbin{} uses a negative binomial distribution, but does not scale inputs and
outputs of the RNN and training instances are drawn uniformly rather than using
weighted sampling.
\end{itemize}

As in \citep{seeger2016}, we use
 $\rho$-risk metrics (quantile loss) that quantify the accuracy of a quantile $\rho$ of the predictive distribution; 
 the exact definition of these metric is given in supplementary material.
The metrics are evaluated for a certain \emph{spans} $[L, L+S)$ in the prediction range,
where $L$ is a \emph{lead time} after the forecast start point.
Table~\ref{tab:accuracy-distribution} shows the $0.5$-risk and $0.9$-risk and for different lead
times and spans. Here all(K) denotes the average risk of the marginals $[L, L+1)$ for $L<K$.
We normalize all reported metrics with respect to the strongest
previously published method (baseline).
\deepar{} performs significantly better than all other methods on these
datasets. The results also show the importance of modeling these data sets with
a count distribution, as \gaussian{} performs significantly worse. The \ec{} and \ecsub{}
data sets exhibit the power law behavior discussed above, and without scaling and weighted
sampling accuracy is decreased (\negbin{}). On the \parts{} data set, which does not exhibit the 
power-law behavior, \negbin{} performs similar to \deepar{}.

\begin{wraptable}{L}{0.4\textwidth}
\centering
\scriptsize
\begin{tabular}{l|rrrrr}
\toprule
{} & \multicolumn{2}{c}{\electricity} & \multicolumn{2}{c}{\traffic} \\
{} & ND & RMSE & ND & RMSE \\
\midrule 
\matfact{} & 0.16 & 1.15 & 0.20 & 0.43 \\
\deepar{} & \bf{0.07} & \bf{1.00} & \bf{0.17} & \bf{0.42} \\
\bottomrule
\end{tabular}
\caption{Comparison with \matfact \label{tab:accuracy-matfact}}
\end{wraptable}

In Table \ref{tab:accuracy-matfact} we compare point forecast accuracy on the \electricity{} and \traffic{} datasets
against the matrix factorization technique (\matfact{}) proposed in \cite{NIPS2016_6160}.
We consider the same metrics namely Normalized Deviation (ND) and Normalized RMSE (NRMSE) whose definition are given
in the supplementary material.
The results show that DeepAR outperforms \matfact{} on both datasets.
%A possible extension to alleviate this issue would be to model multivariate 
%time series of modest dimension with their covariate matrix or use a low-rank approximation 
%in case the dimension is too large.

\subsection{Qualitative analysis}
Figure \ref{fig:predictionsplots} shows example predictions from the \ec{} data set.
In Fig.~\ref{fig:uncertainty}, we show aggregate sums of different quantiles of the
marginal predictive distribution for \deepar{} and \issm{} on the \ec{}
dataset.  In contrast to ISSM models such as \cite{seeger2016}, where a linear
growth of uncertainty is part of the modeling assumptions,
the uncertainty growth pattern is learned from the data.  In this case, the model
does learn an overall growth of uncertainty over time. However, this is not simply
linear growth: uncertainty (correctly) increases during Q4, and decreases again shortly
afterwards.

The calibration of the forecast distribution is depicted in
Fig. \ref{fig:calibration}.  Here we show, for each percentile $p$
the $\text{Coverage}(p)$, which is defined as the fraction of time series in
the dataset for which the $p$-percentile of the predictive distribution is
larger than the the true target.  For a perfectly calibrated prediction
it holds that $\text{Coverage}(p) = p$, which corresponds to the diagonal.
Compared to the \issm{} model, calibration is improved overall.

To assess the effect of modeling correlations in the output, i.e., how much
they differ from independent distributions for each time-point, we plot the
calibration curves for a shuffled forecast, where for each time point
the realizations of the original forecast have been shuffled, destroying any correlation between time steps. 
For the short lead-time span (left) which consists of just one
time-point, this has no impact, because it is just the marginal
distribution. For the longer lead-time span (right), however, destroying the correlation
leads to a worse calibration, showing that important temporal correlations 
are captured between the time steps.

\section{Conclusion}

We have shown that forecasting approaches based on modern deep learning techniques can drastically
improve forecast accuracy over state of the art forecasting methods on a wide
variety of data sets.  
Our proposed DeepAR model effectively learns a \emph{global} model from related time series,
handles widely-varying scales through rescaling and velocity-based sampling,
generates calibrated probabilistic forecasts with
high accuracy, and is able to learn complex patterns such as
seasonality and uncertainty growth over time from the data.

Interestingly, the method works with little or no hyperparameter tuning on a wide variety
of datasets, and in is applicable to medium-size datasets containing only a few hundred time series.

\section{Supplementary materials}

\subsection*{Error metrics}

\subsubsection*{$\rho$-risk metric}

The aggregated target value of an item $i$ in a span is denoted as $Z_i(L,S) = \sum_{t=t_0+L}^{t_0 + L + S}
z_{i,t}$. For a given quantile $\rho\in(0,1)$ we denote the predicted
$\rho$-quantile for $Z_i(L,S)$ by $\hat{Z}^\rho_i(L, S)$. To obtain such
a quantile prediction from a set of sample paths, each realization is first
summed in the given span. The samples of these sums then represent the
estimated distribution for $Z_i(L,S)$ and we can take the $\rho$-quantile
from the empirical distribution.

The $\rho$-quantile loss is then defined as
\begin{equation*}
    L_{\rho}(Z, \hat{Z}^\rho) = 2 (\hat{Z} - Z)\left(\rho \mathrm{I}_{\hat{Z}^\rho>Z} - (1 - \rho) \mathrm{I}_{\hat{Z}^\rho \leq Z}\right) \;.
\end{equation*}
In order to summarize the quantile losses for a given span across all items, we consider a normalized sum of quantile
losses $\left(\sum_i L_\rho(Z_i, \hat{Z}_i^\rho\right) / \left(\sum_i Z_i\right)$, which we call the $\rho$-risk.  

\subsubsection*{ND and RMSE metrics}
ND and RMSE metrics are defined as follow:
\begin{align*}
 \text{ND} &= \frac{\sum_{i,t} |z_{i, t} - \hat{z}_{i, t}|}{\sum_{i,t} |z_{i, t}|} & 
 \text{RMSE} &= \frac{\sqrt{\frac{1}{N (T - t_0)}\sum_{i,t}(z_{i,t} - \hat{z}_{i, t})^2}}{\frac{1}{N (T - t_0)}\sum_{i,t} |z_{i,t}|}
\end{align*}
where $\hat{z}_{i,t}$ is the predicted median value for item $i$ at time $t$ and the sums are over all items and all time points in the
prediction period.

\subsection*{Experiment details}

We use MxNet as our neural network framework \cite{mxnet}. Experiments are run on a laptop for \parts{} and with a single AWS p2.xlarge instance (four core machine with a single GPU) for other datasets. Note that predictions can be done in all datasets end to end in a matter of hours even with a single machine.
We use ADAM optimizer \cite{adam} with early stopping and standard LSTM cells with a forget bias set to $1.0$ in all experiment and 200 samples are drawn from our decoder to generate predictions.

\begin{table*}[htb]
\footnotesize
\centering
\begin{tabular}{cccccc}
\midrule
 & \parts & \electricity & \traffic & \ecsub & \ec \\\midrule
$\#$ time series & 1046 & 370 & 963 & 39700 & 534884  \\
time granularity & month & hourly & hourly & week & week \\
%$\#$ training time observations& 50K & 26K & 10K & 6M & 22M & 20M \\
domain& $\mathbb{N}$ & $\mathbb{R}^+$ & $[0,1]$ &  $\mathbb{N}$ & $\mathbb{N}$ \\
encoder length & 8 & 168 & 168 & 52 & 52 \\
decoder length & 8 & 24 & 24 & 52 & 52 \\
$\#$ training examples & 35K & 500K & 500K & 2M & 2M \\
item input embedding dimension & 1046 & 370 & 963 & 5 & 5 \\
item output embedding dimension & 1 & 20 & 20 & 20 & 20 \\
batch size & 64 & 64 & 64 & 512 & 512 \\
learning rate & 1e-3 & 1e-3 & 1e-3 & 5e-3 & 5e-3  \\
$\#$ LSTM layers & 3 & 3 & 3 & 3 & 3 \\
$\#$ LSTM nodes & 40 & 40 & 40 & 120 & 120 \\
running time & 5min & 7h & 3h & 3h & 10h \\
\midrule
\end{tabular}
\caption{Datasets statistics and RNN parameters \label{tab:datasets}}
\end{table*}

For \parts{} dataset, we use the 42 first months as training data and report error on the remaining 8. For the other datasets \electricity{}, \traffic{}, \ecsub{} and \ec{} the set of possible training instances is sub-sampled to the number indicated in table \ref{tab:datasets}. The scores of \electricity{} and \traffic{} are reported using the rolling window operation described in \cite{NIPS2016_6160}, note that we do not retrain our model but reuse the same one for predicting across the different time windows instead. 
Running times measures an end to end evaluation, e.g. processing features, training the neural network, drawing samples and evaluating produced distributions.

For each dataset, a grid-search is used to find the best value for the hyper-parameters \emph{item output embedding dimension} and \emph{\# LSTM nodes} (e.g. hidden number of units). To do so, the data before the forecast start time is used as training set and split into two partitions.
For each hyper-parameter candidate, we fit our model on the first partition of the training set containing 90\% of the data and we pick the one 
that has the minimal negative log-likelihood on the remaining 10\%.
Once the best set of hyper-parameters is found, the evaluation metrics (0.5-risk, 0.9-risk, ND and RMSE) are then evaluated on the test set, e.g. 
the data coming after the forecast start time. 
Note that this procedure could lead to over-fitting the hyper-parameters to the training set but this would then also degrades the metric we report. 
A better procedure would be to fit parameters and evaluate negative log-likelihood not only on different windows but also on non-overlapping time intervals. 
As for the learning rate, it is tuned manually for every dataset and is kept fixed in hyper-parameter tuning.
Other parameters such as encoder length, decoder length and item input embedding are considered domain dependent
and are not fitted. Batch size is increased on larger datasets to benefit more from GPU's parallelization. Finally, running times measures an end to end evaluation, e.g. processing features, training the neural network, drawing samples and evaluating produced distributions.

%%%% Perhaps we just leave this figure away.
%
% \begin{figure}
% \centering
% \includegraphics[width=0.4\textwidth]{plots/scaling_vs_noscaling.pdf}
%     \caption{RMSE of the forecast with velocity scaling (y-axis) and without
%     velocity scaling (x-axis) in logarithmic scale.  The solid straight line
%     shows a linear fit of the data.  The tilting with respect to the diagonal
%     (dashed line) indicates the increased error for higher velocities. 
%     \label{fig:scaling}}
% \end{figure}
% The effect of our scaling correction is show in Fig.~\ref{fig:scaling}.
% This scatter plot depicts the RMSE in log scale with and without scaling correction.
% Without scaling, faster items exhibit larger error than slower items which is
% highly problematic for demand forecasting. The difficulty for the neural
% network to learn from few high velocity examples also results in a higher
% overall error if the scaling correction is omitted.

\subsection*{Missing Observations}
In some forecasting settings, the target values $z_{i,t}$ might be missing (or unobserved) for a subset of the time points.
For instance, in the context of demand forecasting, an item may be
out-of-stock at a certain time, in which case the demand for the item cannot be observed.
Not explicitly modeling such missing observations (e.g.\ by assuming that the observed sales
correspond to the demand even when an item is out of stock), can, in the best case, lead to systematic forecast underbias,
and, in a worst case in the larger supply chain context, can lead to a disastrous downward spiral where an out-of-stock situation 
leads to a lower demand forecast, lower re-ordering and more out-of-stock-situations.
In our model, missing observations can easily be handled in a principled way by replacing each unobserved value $z_{i,t}$ by a sample
$\tilde{z}_{i,t} \sim \ell(\cdot|\theta(\hVec_{i,t}))$ from the conditional predictive distribution when computing (1), 
and excluding the likelihood term corresponding to the missing observation from (2).%
We omitted experimental results in this setting from the paper, as doing a proper evaluation in the light of missing data in the prediction range requires non-standard adjusted metrics that are hard to compare across studies (see e.g.\ \citep{seeger2016}).

%There are many future directions for extensions and improvements.  
%Problem specific likelihoods such as zero inflated negative binomial likelihood are easy
%to incorporate.  
%An interesting direction would be to not postulate a fixed likelihood, but work
%with other generative models, that learn a distribution from the data such as \cite{gan}.
%\todo{As shown in \traffic{} example, our model need to have independent time series to work well. 
%Modeling dependent time series either as multivariate with covariance matrix or via a hierarchy would 
%another}
%}
% not very incisive I think
%Sequence-to-sequence approaches can be used when the available covariates differ between conditioning and prediction period.
% Acknowledgements should only appear in the accepted version.
%\section*{Acknowledgements}
%The authors wish to thank the Lutter team and in particular Tim Januschowski for his continous %support and faith in the project.
%\clearpage
\bibliographystyle{plainnat}
\bibliography{deepar-arxiv.bib}

%%%% Perhaps we just leave this figure away.
%
% \begin{figure}
% \centering
% \includegraphics[width=0.4\textwidth]{plots/scaling_vs_noscaling.pdf}
%     \caption{RMSE of the forecast with velocity scaling (y-axis) and without
%     velocity scaling (x-axis) in logarithmic scale.  The solid straight line
%     shows a linear fit of the data.  The tilting with respect to the diagonal
%     (dashed line) indicates the increased error for higher velocities. 
%     \label{fig:scaling}}
% \end{figure}
% The effect of our scaling correction is show in Fig.~\ref{fig:scaling}.
% This scatter plot depicts the RMSE in log scale with and without scaling correction.
% Without scaling, faster items exhibit larger error than slower items which is
% highly problematic for demand forecasting. The difficulty for the neural
% network to learn from few high velocity examples also results in a higher
% overall error if the scaling correction is omitted.
\end{document}